# The Indus Script and Economics.
# A Role for Indus Seals and Tablets in Rationing and Administration of Labor

Rajesh P. N. Rao

The Indus script remains one of the last major undeciphered scripts of the ancient world. We focus here on Indus inscriptions on a group of miniature tablets discovered by Meadow and Kenoyer in Harappa in 1997. By drawing parallels with proto-Elamite and proto-Cuneiform inscriptions, we explore how these miniature tablets may have been used to record rations allocated to porters or laborers. We then show that similar inscriptions are found on stamp seals, leading to the potentially provocative conclusion that rather than simply indicating ownership of property, Indus seals may have been used for generating tokens, tablets and sealings for repetitive economic transactions such as rations and exchange of canonical amounts of goods, grains, animals, and labor in a barter-based economy.

In an interesting article, Richard H. Meadow and Jonathan Mark Kenoyer describe their discovery of a group of 22 three-sided 'tiny steatite seals' containing identical inscriptions in the Indus script (Meadow and Kenoyer 2000) (Figure 1). The inscription on each side consists of two symbols, one of which appears to be a numeral. In this article, we investigate the hypothesis that these miniature tablets may have been used as ration tokens for rations allocated to porters or laborers. Such a practice would be in keeping with bookkeeping practices in the 3rd millennium BC in the contemporaneous Mesopotamian and Elamite civilizations with whom the Indus civilization had active trading relationships.

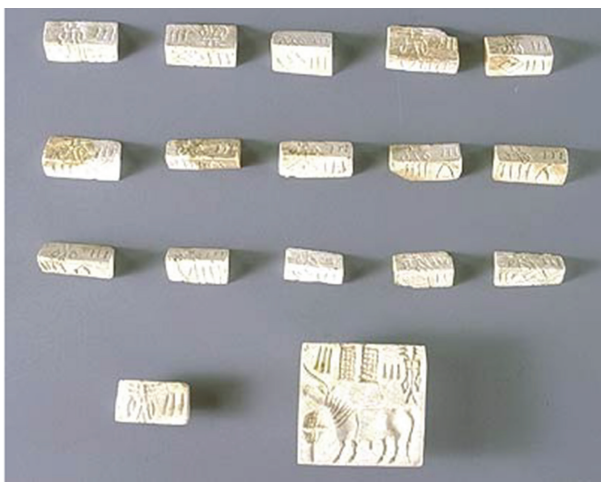

Figure 1. Sixteen three-sided miniature tablets and a 'unicorn' seal discovered at Harappa. Each side of each tablet has a short incised inscription in the Indus script. The 16 tablets were found together in a clump dumped over a wall. As noted by Meadow and Kenoyer, the inscription on one of the three sides of the tablets is identical to the last two signs on the seal which was discovered in a nearby area (image from www.harappa.com, courtesy Harappa Archaeological Research Project).

The insights from these miniature tablets and the fact that very similar inscriptions are found on stamp seals point to the potentially provocative conclusion that rather than containing names of owners as traditionally believed, a vast majority of Indus seals may have been used to generate tokens, tablets, and sealings for repetitive economic transactions such as rations and exchange of canonical amounts of goods, grains, animals, and labor in a barter-based economy. Following Wells (2015) and Bonta (2010), we discuss a partitioning of the corpus of Indus inscriptions into 'Patterned Texts', tailored for economic transactions, and 'Complex Texts', such as the Dholavira signboard (Figure 2), which may encode other types of linguistic information. We then draw parallels between the components of patterned Indus texts and proto-Elamite texts recording economic transactions.

### The Indus Script

More than 4700 inscriptions in the Indus script have been unearthed on stamp seals, sealings, copper tablets, copper tools, ivory rods, pottery, and miniature tablets. A presumed 'signboard' has also been discovered in Dholavira (Figure 2). Despite the large number of claimed decipherments (see Possehl 1996 for a review), none has been widely accepted by the community. Obstacles to decipherment include the lack of bilinguals, the brevity of the inscriptions, and our lack of knowledge of the language(s) used in the civilization. A prerequisite to decipherment is identifying the basic signs in script. After analysis of the positional statistics of signs in the corpus of known inscriptions (c. 1977), Mahadevan arrived at a list of 417 independent signs in his concordance (Mahadevan 1977). Parpola used similar methods to estimate a slightly shorter list of 386 signs (Parpola 1994) while Wells has more recently proposed a list of 694 signs (Wells 2015). Figure 3 shows a subset of signs from Mahadevan's list.

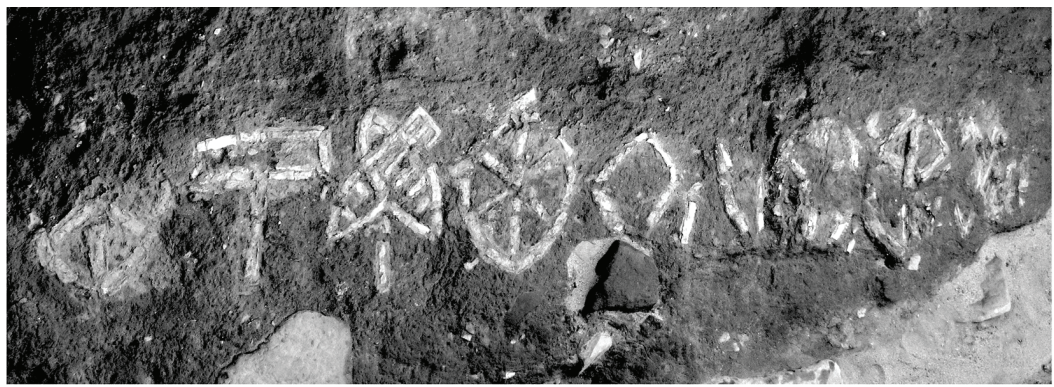

Figure 2. Dholavira signboard (photograph by the author).

Although the script remains undeciphered, there is widespread consensus on the direction of writing in the script. Barring a few exceptions, the writing direction is predominantly right to left (i.e., left to right in seals and right to left in the impressions). There exists convincing external and internal evidence supporting this claim (e.g., Mahadevan 1977; Parpola 1994).

In our previous work, we used Markov models to gain a quantitative understanding of the syntactic patterns in the Indus texts. Specifically, we computed the probabilities of symbols following each other (Rao et al. 2009b) and the probabilities of different length sequences (Yadav et al. 2010). We also compared the 'flexibility' of Indus texts sequences of different lengths, as quantified by entropy, to the entropies of other scripts and nonlinguistic sequences. We found that the Indus texts show a similar amount of 'flexibility' as other scripts (Rao 2010; Rao et al. 2009a) (Figure 4). These results, while not proving that the Indus script is a full-fledged writing system, provide evidence that the rules of the script allow a similar degree of flexibility in symbol combinations as other linguistic scripts.

Our previous analyses of the script were conducted over the entire corpus of Indus inscriptions and did not make any distinctions between inscriptions on different types of objects. To begin the process of understanding the function and use of the Indus script, we now examine the inscriptions on a very restricted group of objects, namely, the set of 16 miniature tablets discovered by Meadow and Kenoyer at Harappa (Figure 1). We try to understand the purpose of these tablets in the context of ancient bookkeeping and accounting.

## Harappan Miniature Tablets as Ration Tokens: Parallels with Near Eastern Economic Practices

The inscriptions on the 16 Harappan miniature tablets are depicted more clearly in Figure 5 and are as follows: ⚭|||, ◇||| and ∪||||.

The last inscription ∪|||| is of particular interest because similar inscriptions have been found on pots. By estimating the approximate volume of one of these pots, Wells has proposed that the symbol ∪ is a volumetric measure with a value of approximately 40 liters (Wells 2015). If this is correct, the value of ∪|||| would be approximately 160 liters.

We do not know what the symbols ⚭ and ◇ stand for, but assuming the three strokes in front of each symbol denote the numeral three, a plausible hypothesis is that these symbols stand for measures or nouns that are being counted (Wells 2015), similar to the volumetric measure ∪.

At this point, it is beneficial to consider similar types of proto-Elamite and proto-Cuneiform inscriptions that were utilized by Near Eastern civilizations with whom the Indus civilization had trade contacts. Figure 6 shows examples of proto-Elamite and proto-Cuneiform tablets with a similar syntactic structure as the Harappan miniature tablets, namely, brief two-symbol inscriptions, each consisting of a numeral followed by a measure or noun. Many of these inscriptions list rations of barley, beer, and oil given to carters, porters, couriers, or laborers. Note the similarity in counting measures based on a rationing vessel in Figure 6 and the Harappan inscription ∪|||| counting the volumetric measure given by the vessel-like symbol ∪.

There is considerable archaeological evidence that merchants in Indus cities such as Harappa traded extensively with other cities of the Indus civilization (Kenoyer 1998). Goods from farmers, food and craft manufacturers, potters, metalsmiths, timber merchants and others needed to be transported by porters and carters. Similar to the rationing practices of their Near Eastern counterparts, merchants in Indus cities most likely employed some type of rationing system to compensate their employees who were porters, carters or other types of laborers.

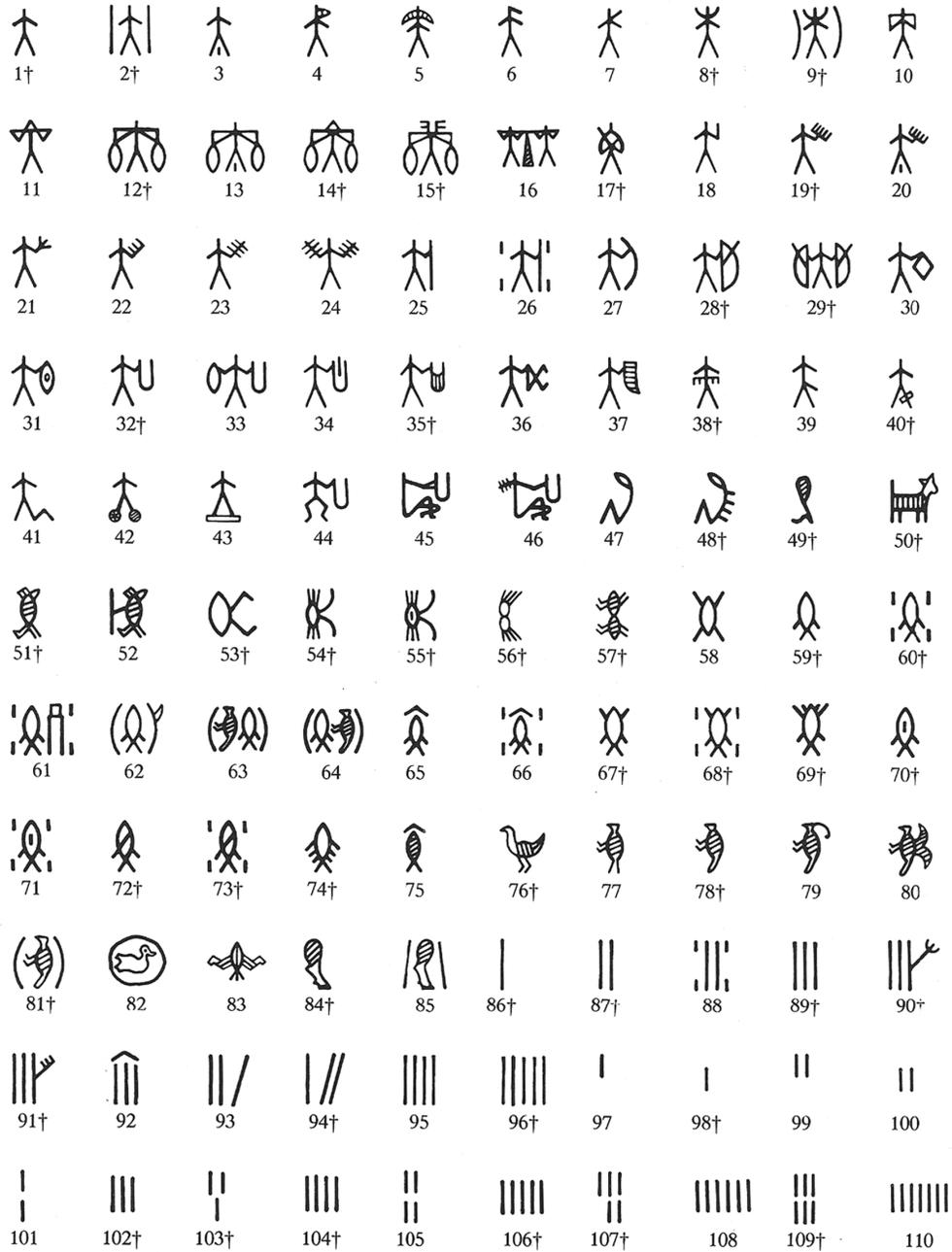

Figure 3. First 110 Indus signs from Mahadevan's list of 417 signs (from Mahadevan 1977).

We therefore propose that the 16 Harappan miniature tablets were part of a rationing system used by a merchant family or business establishment for a group of porters employed to transport goods. This suggestion is based on the fact that the symbol ⚍ clearly depicts a person carrying two bundles of goods tied to a pole carried across the shoulders. Persons using this same technique for carrying goods can still be occasionally spotted in rural India today. By analogy with proto-Elamite and proto-Cuneiform tablets, the inscription ∪ ||| may denote four measures of barley or other grain, while the inscription ◇ ||| may denote three measures of beer, oil, or another commodity paid in compensation for transport of goods. The numeral ||| quantifying ⚍ may denote three measures of a particular kind of porter-

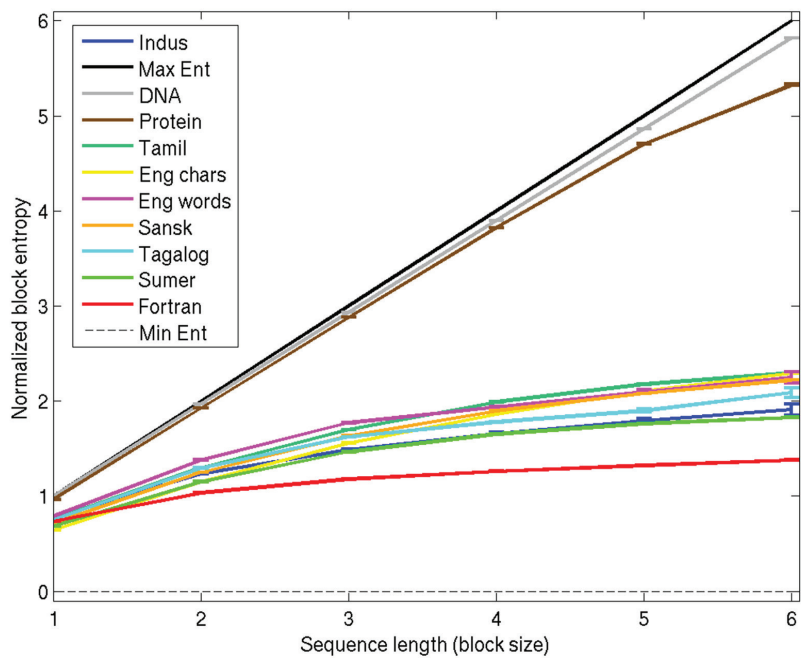

Figure 4. Entropy of the Indus script compared to natural languages and other sequences. Symbols were signs for the Indus script, bases for DNA, amino acids for proteins, characters for English, words for English, Tagalog and Fortran, symbols in abugida (alphasyllabic) scripts for Tamil and Sanskrit, and symbols in the cuneiform script for Sumerian (for details regarding these datasets, see Rao *et al.* 2009a). To compare sequences over different alphabet sizes L, the logarithm in the entropy calculation was taken to base L (417 for Indus, 4 for DNA, etc.). The resulting normalized block entropy is plotted as a function of block size. Error bars denote 1 standard deviation above/below mean entropy and are negligibly small except for block size 6 (from Rao 2010).

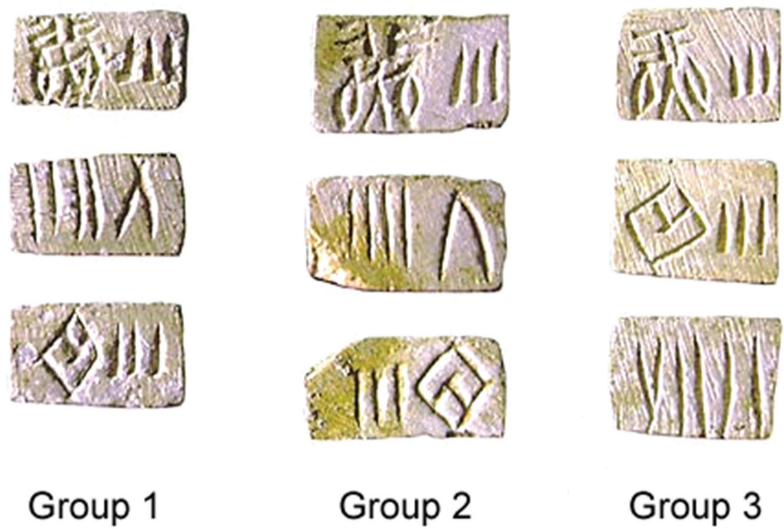

Figure 5. Inscriptions on the three sides of three of the sixteen miniature tablets show in Figure 1. These three tablets are representative of three different groups identified by Kenoyer as originating from the hands of at least three different stone engravers (image from www.harappa.com, courtesy Harappa Archaeological Research Project).

type labor. Note that the symbol 🕱 can be decomposed as ∪+ 🕱. The ligature of the most commonly occurring Indus symbol ∪ with the unadorned porter symbol 🕱 may signify a particular kind of porter, in contrast to the symbol 🕱 (= ↑+ 🕱) which may denote a different type of porter. Similarly, other symbols potentially related to measures of porter-style work (from Mahadevan 1977) include: 🕱, 🕱, and 🕱.

The identification of the above anthropomorphic symbols with 'porter'-type labor immediately suggests similar labor-related functions for some of the other anthropomorphic signs:

🕱 : man + wheels = carter-type labor?
🕱 : man + harrow = farm labor?
🕱 : man + tongs = metalsmithing labor?
🕱 : man + bow/arrow = hunting-related labor?

Without additional external evidence, the above suggestions about various measures of labor must be considered speculative but we hope these suggestions will open up new lines of inquiry regarding the anthropomorphic signs different from previous suggestions (Mahadevan 1975), with an emphasis on their possible role in rationing and exchanging goods in compensation for labor by employees of Indus institutions and business establishments.

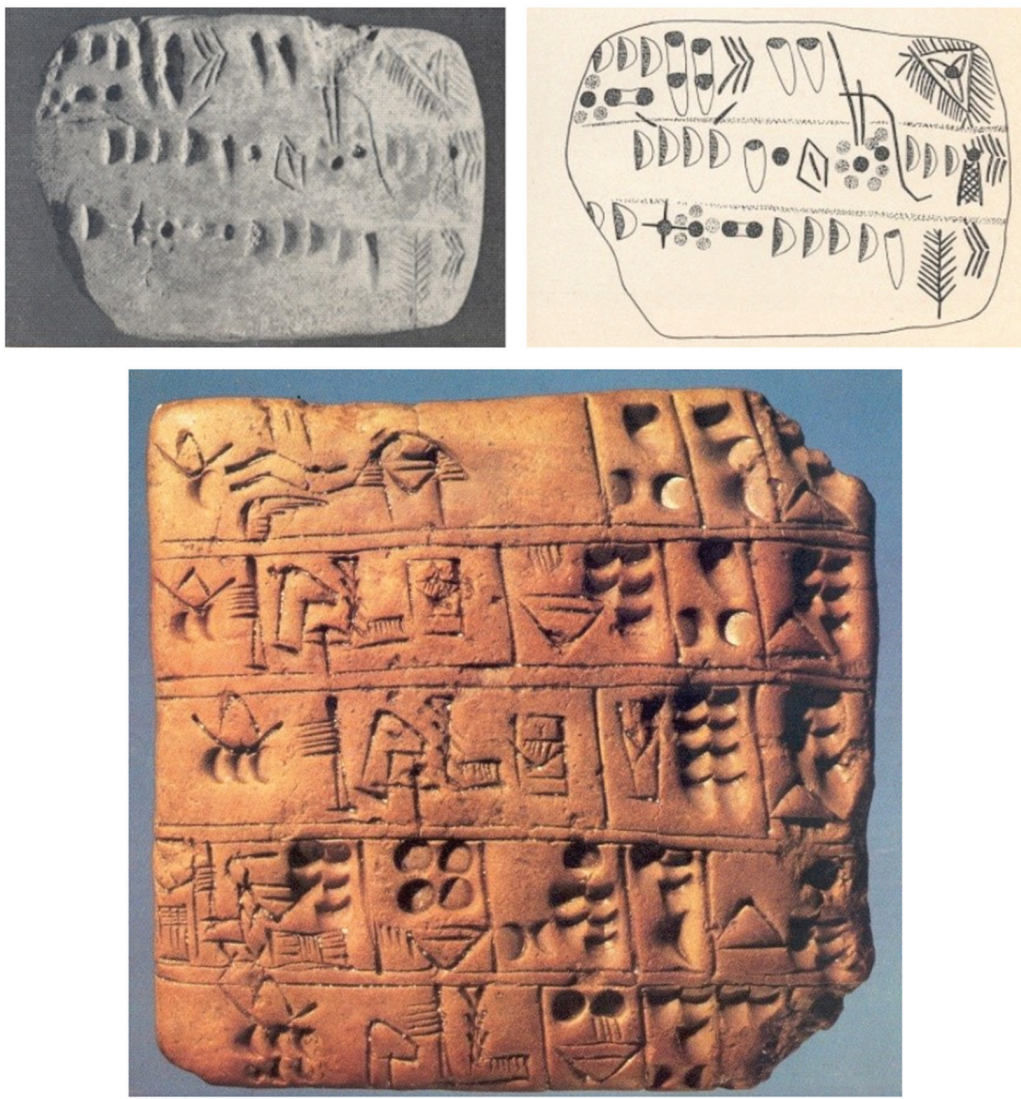

Figure 6. A Proto-Elamite tablet and a Proto-Cuneiform tablet: Proto-Elamite tablet counting measures of cereal and grain (with an illustration of the tablet on the right). The 'hairy triangle' is thought to denote an issuing institution while the commodities being counted are represented by symbols such as the stalk of barley (second from right at the bottom) (top); Proto-Cuneiform tablet enumerating the rations for workers over a five-day week. Note the numerous instances of numerals in front of a rationing vessel (a beveled-rim bowl) (bottom, courtesy Trustees of the British Museum).

**Implications of the Mode of Disposal of the Miniature Tablets**

Meadow and Kenoyer note that the 16 miniature tablets were found as a cluster outside a perimeter wall, strongly suggesting the entire group was intentionally disposed. Why did the establishment possessing these tablets intentionally discard them? One plausible explanation is that these tablets were no longer valid 'currency' for rations or exchange of goods for labor. We hypothesize that an alternate system, one that is harder to forge and based on the legitimized authority of stamp seals with their iconic (and potentially totemic) imagery (such as the zebu bull or 'unicorn') and characteristic of particular business establishments may have come into vogue. This suggests a new role for the ubiquitous Indus stamp seal in the administration of labor and exchange of goods in a barter-based economy in the Indus valley (Kenoyer 1998).

**From Miniature Tablets to Indus Seals**

Figure 7B illustrates a two-sided seal from Mohenjo-daro. On one side, the seal produces the inscription "◊∪∪ when stamped – this stamped inscription shares similarities with the two inscriptions ◊||| and ∪|||| on the Harappan miniature tablets (Figure 7A). The second side of the seal produces the two-symbol inscription which is similar to the inscription on the miniature tablet.

We propose that rather than inscribing by hand a large number of identical tablets (as in the case of the Harappan miniature tablets), stamp seals such as those shown in Figures 7C and 7D may have been introduced as an efficient way for business establishments to generate on demand a large number of ration tokens or wage tokens for hired labor, and tokens for facilitating goods and labor exchange in a barter-based economy. A striking example of the use of stamp seals for such a purpose may be the three pendant-like tokens (Figure 7E) discovered by a team of Japanese researchers at Kanmer in Kutch, India (Kharakwal *et al.* 2013). These tokens all have identical seal impressions from a 'unicorn' stamp seal. All three have holes in the center possibly for stringing together several such tokens, and a two-symbol inscription consisting of a single long stroke (likely the numeral 1) and an anthropomorphic symbol which we labeled 'man + harrow' above. We hypothesize that this anthropomorphic symbol may represent a form of labor such as farm labor, similar to the 'porter' symbol on the Harappan miniature tablets (Figure 7A).

**Indus seals and their possible role in a barter-style economy**

Our hypothesis that the Indus seals were used to facilitate exchange of goods and labor in a barter-style economy raises the following question: can the longer inscriptions found on Indus seals be interpreted in light of this hypothesis? A number of authors have previously suggested 'grammatical' rules for explaining the structure of Indus inscriptions (Bonta 2010; Mahadevan 1986; Parpola 1994, Wells 2012, 2015). Following Wells and Bonta, we partition the corpus of Indus texts into two sets: Patterned Texts, which tend to be stereotypical and can be defined by the rules below, and Complex Texts, which may contain linguistic constructs not captured by these rules (e.g., the Dholavira 'signboard' in Figure 2, which does not appear to contain any numeral symbols).

We characterize Patterned Texts here using a 'grammar' or a set of rules for generating the strings of Indus symbols that constitute these texts (→ denotes 'generates' and | denotes 'or'):

Patterned → Terminal Core Medial Prefix
Prefix → | | | | | | etc.
Medial → Fish-Oval-Cluster | Fish-Numeral-Cluster | etc.
Fish-Oval-Cluster → | | | etc.
Fish-Numeral-Cluster → | | etc.
Core → | | | etc.
Terminal → | | | | | etc.

The rules above are meant to illustrate the general structure of Patterned Texts – the reader is referred to the work of the authors cited above for various nuances and an in-depth treatment. For the purpose of this article, it is sufficient to consider the four-part segmentation of patterned Indus texts to draw parallels with proto-Elamite inscriptions.

Indus patterned text format (rewritten in left-to-right format):
*Prefix Medial Core Terminal*
Proto-Elamite inscription format on rationing tablets:
*Heading (function of tablet) Person/Institution Commodity Number*

Based on the above comparison, we suggest the following interpretation of a four-part patterned Indus text:
*Prefix*: Institution/Business/Landlord/Family/Person
*Medial*: Number/Quantity/Measures
*Core*: Commodity
*Terminal*: Function of sealing/tablet

The above attributions of function to the components of Indus patterned texts are motivated by the following considerations. The 'Medial' component in Indus texts typically involves numbers and 'fish' signs. The latter have been linked by Bonta to the ancient weight measures of 'minas' or 'maashas' (Bonta 2010). The 'Core' component follows the numerical component in Indus texts and therefore, following the format of the miniature tablets, is a prime candidate for a noun, object or commodity being counted – this would include strings such as , , and , which may name different commodities. The 'Prefix' component includes the commonly occurring sign ◊ as well as longer strings such as . We hypothesize that the Prefix may denote a person, business, landlord, family, or institution (cf. Mahadevan's attempt to link the symbol ◊ to the citadel or principal quarter of the city, see Mahadevan 2010). Note that the use of ◊ as a Prefix in a patterned text suggests a polyvalent use of this symbol if, as we suggested earlier, it also denotes a commodity that can be counted, as in the inscription ◊||| on miniature tablets.

Finally, the 'Terminal' component may indicate the overall function of the token, sealing or tablet with the seal's impression, e.g., whether the token indicates rations owed to a person or issued by an institution, whether it is a receipt for goods received, or whether it is a sealing indicating the quantity/quantities and commodity/commodities inside a bundle of goods. The idea that the Terminal component of patterned Indus texts might capture the function of a particular seal is consistent with attempts by Bonta (2010) to equate terminal symbols such as ∪ and ↑ with predicates 'be, have, own' and 'be worth, be valued' respectively, which help to clarify the role of the token or tablet in an economic transaction.

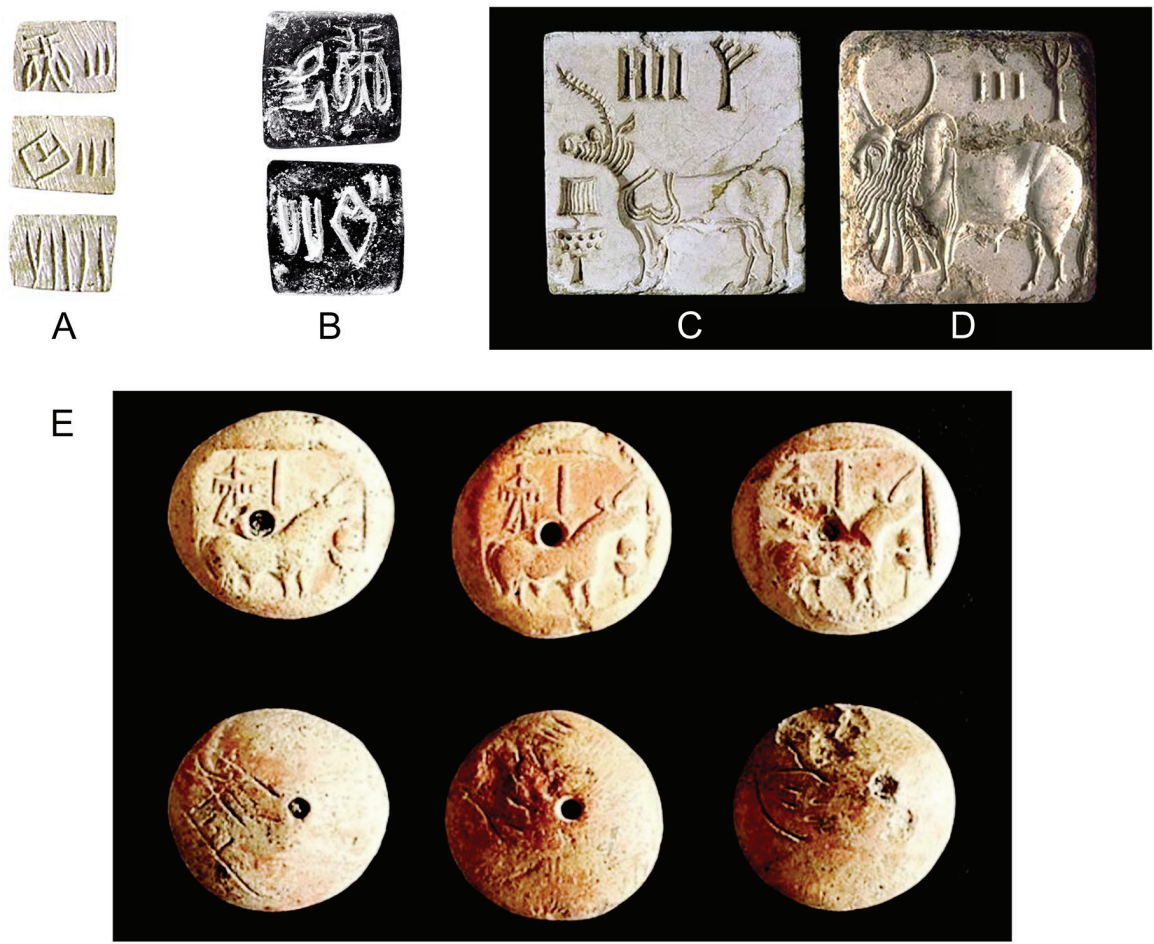

Figure 7. (A) Three sides of a Harappan miniature tablet; (B) two-sided seal from Mohenjo-daro with inscriptions similar to the Harappan miniature tablets; (C) and (D) Two typical Indus stamp seals, with 'unicorn' and zebu bull motifs and a 'numeral + grain stalk' inscription counting a grain-related commodity. Compare with the proto-Elamite 'barley stalk + numeral' inscription in Figure 6; (E) Tokens with identical seal impressions and holes in the center, unearthed at Kanmer in Kutch, India (A-D, from www.harappa.com, courtesy Harappa Archaeological Research Project; E, courtesy Toshiki Osada).

## Conclusions

Starting with the hypothesis that the cluster of miniature tablets discovered by Meadow and Kenoyer may have been used as ration tokens for porters, we proposed that Indus seals may have been invented not to simply indicate ownership of property, as in other ancient civilizations, but to efficiently generate large numbers of economic tokens, tablets and sealings for repetitive transactions.

Our proposal is supported by several lines of archaeological evidence: (1) no Indus seal has ever been found in a burial in the Indus valley, unlike other ancient societies where seals were used to indicate ownership and were buried with the owner, (2) many seals appear to have been deliberately broken and discarded, in keeping with the hypothesis that they may have outlived their usefulness for particular economic transactions, (3) tokens with seal impressions such as those in Figure 7E have holes in them suggesting they were strung together and worn or carried by a laborer or by an administrator, (4) stamp seals themselves have a boss on the back with a hole for possibly stringing together multiple seals, opening up the possibility that an individual may have carried multiple seals and selected the appropriate one for generating a token or sealing according to the economic transaction at hand. Additional analysis and archaeological evidence are needed to test our ascription of specific economic functions to the components of patterned Indus texts based on proto-Elamite parallels.

**Acknowledgments**

This work was supported by a Guggenheim Fellowship to RPNR.

**Bibliography**

Bonta, S. *The Indus Valley Script: A New Interpretation. URL:* http://www.academia.edu/8691466/The_Indus_Valley_Script_A_New_Interpretation

Kharakwal, J. S., Rawat, Y. S. and Osada, T. 2013. Excavation at Kanmer, a Harappan site in Gujarat: some observations. In D. Frenez and M. Tosi (eds), *South Asian archaeology 2007* (British Archaeological Reports 2454). Oxford, Archaeopress.

Kenoyer, J. M. 1998. *Ancient cities of the Indus Valley Civilisation.* Oxford, Oxford University Press.

Mahadevan, I. 1979. *Study of the Indus script through bilingual parallels. In G. L. Possehl (ed.), Ancient Cities of the Indus:* 261-267. New Delhi.

Mahadevan, I. 1977. *The Indus Script. Texts, Concordance and Tables* (Memoirs of the Archaeological Survey of India, 77). New Delhi, Archaeological Survey of India.

Mahadevan, I. 1986. Towards a grammar of the Indus texts: 'intelligible to the eye, if not to the ears'. *Tamil Civilization* 4(3-4): 15-30.

Mahadevan, I. 2010. *Akam and Puram: 'Address' Signs of the Indus Script. URL:* http://rmrl.in/?page_id=1044

Meadow, R. H. and Kenoyer, J. M. 2000. The 'Tiny Steatite Seals' (Incised Steatite Tablets) of Harappa: Some Observations on Their Context and Dating. In M. Taddei and G. De Marco (eds), *South Asian Archaeology 1997, Volume 1:* 321–340. Rome, Istituto Italiano per l'Africa e l'Oriente.

Parpola, A. 1994. *Deciphering the Indus script.* Cambridge, Cambridge University Press.

Possehl, G. L. 1996. *The Indus Age. The Writing System.* Philadelphia, PA, University of Pennsylvania Press.

Rao, R. P. N. 2010. Probabilistic Analysis of an Ancient Undeciphered Script. *IEEE Computer* 43(4): 76-80.

Rao, R. P. N., Yadav, N., Vahia, M. N., Joglekar, H., Adhikari, R. and Mahadevan, I. 2009a. Entropic evidence for linguistic structure in the Indus script. *Science* 324(5931): 1165. DOI: 10.1126/science.1170391.

Rao, R. P. N., Yadav, N., Vahia, M. N., Joglekar, H., Adhikari, R. and Mahadevan, I. 2009b. A Markov model of the Indus script. *Proceedings of the National Academy of Sciences* 106: 13685-13690.

Wells, B. K. 2012. *Epigraphic Approaches to Indus Writing.* Cambridge, MA, Oxbow Press.

Wells, B. K. 2015. *The Archaeology and Epigraphy of Indus Writing.* Oxford, Archaeopress.

Yadav, N., Joglekar, H., Rao, R. P. N., Vahia M. N., Adhikari R. and Mahadevan I. 2010. Statistical analysis of the Indus script using n-grams. *PLoS One* 5(3):e9506.